\documentclass{llncs}

\usepackage{epsfig}

\begin{document}

\newcommand{\hyphtt}[1]{\texttt{\hyphenchar\font45\relax #1}}

\title{Actin - Technical Report}

\author{Raihan H. Kibria}

\institute{Computer Systems Lab\\
Dept. of Electrical Engineering and Information Technology\\
Darmstadt University of Technology,
D-64283 Darmstadt, Germany\\
\email{kibria@rs.tu-darmstadt.de}\\
\texttt{http://www.rs.e-technik.tu-darmstadt.de/}
}

\maketitle

\begin{abstract}
The Boolean satisfiability problem (SAT) can be solved efficiently with variants of the DPLL algorithm.
For industrial SAT problems, DPLL with conflict analysis dependent dynamic decision heuristics has proved to be particularly efficient, e.g.\ in \textsc{Chaff}.
In this work, algorithms that initialize the variable activity values in the solver \textsc{MiniSAT v1.14} by analyzing the CNF are evolved using genetic programming (GP), with the goal to reduce the total number of conflicts of the search and the solving time.
The effect of using initial activities other than zero is examined by initializing with random numbers.
The possibility of countering the detrimental effects of reordering the CNF with improved initialization is investigated.
The best result found (with validation testing on further problems) was used in the solver \textsc{Actin}, which was submitted to SAT-Race 2006.
\end{abstract}

\section{\label{SAT}SAT}

The SAT problem is the question if there exists an assignment to the variables of a Boolean function $f$ so that $f$ evaluates to \textit{true} ($f$ is \emph{satisfiable}), or if no such assignment exists, i.e.\ \textit{f = false} ($f$ is \emph{unsatisfiable}).
SAT is NP-complete.

SAT problems are usually given in \emph{conjunctive normal form} (CNF), consisting of the conjunction of \emph{clauses}, which are disjunctions of \emph{literals} (variables or negated variables).

\subsection{The DPLL Algorithm}

The \emph{Davis(--Putnam)--Loveland--Logemann} algorithm (\emph{DPLL} or \emph{DLL}) \cite{Davis:1962} for SAT operates on Boolean formulas in CNF.
To satisfy a CNF, each clause must be satisfied (i.e.\ contain at least one literal which evaluates to \emph{true}).
\emph{Unit-literal clauses} containing only one literal can only be satisfied if their literal evaluates to \emph{true}; this is a forced assignment or \emph{implication}.
Assigning implications until no further implications are present is called \emph{Boolean constraint propagation (BCP)}.
DPLL searches all variable assignments depth-first for a satisfying set of assignments, applying BCP after making \emph{decision} assignments and \emph{backtracking} when a clause becomes unsatisfied (i.e.\ contains only \emph{false} literals); the latter case is called a \emph{conflict}.
Before the search, BCP is applied on the original CNF.

Classic DPLL has been extended with features such as \emph{non-chronological backtracking}  and \emph{clause learning} \cite{Silva:1996}.
These enabled new decision heuristics which guide the search dynamically by examining learned clauses, e.g.\ \textsc{Chaff}'s \cite{Malik:2001} \emph{Variable State Independent Decaying Sum (VSIDS)} heuristic.
\textsc{MiniSAT} \cite{Een:2005} uses an improved variant of VSIDS.
Each variable has an \emph{activity} associated with it, which is a double-precision floating-point value initialized with 0 (in VSIDS, each literal has its own activity; their initial values are the literal counts in the original CNF).
When a decision has to be made, the variable with the highest activity value is chosen (ties are broken randomly).

After each conflict, an increment value is added to the activities of the variables occurring in the conflict clause, and the increment value is multiplied with a constant greater than 1.
Activities decay $5 \%$ per conflict.
This ensures that recently learned clauses have more influence on the activities.
The activities have to be rescaled once in a while to prevent overflow.
With a small probability, \textsc{MiniSAT} sometimes chooses a random variable.
Decision variables are always assigned the value \emph{false} first.

\subsection{\label{random.init.seq}Initialization with Random Values}

The initial activity values used in \textsc{Chaff} (literal counts) and \textsc{MiniSAT} (all zero) are special cases which might not be ideal, at least for SAT problems derived from industrial hardware verification, e.g.\ bounded model checking (BMC), which will be the focus of this work.
To estimate the effect that the initialization can have, the source code of \textsc{MiniSAT v1.14} was modified to use a random number from a certain range as the initial activity for each variable.
Many different, random initializations could then be compared to the standard in regard to how long a problem takes to solve.
Since measuring the solving time is always imprecise, especially for very short times, the number of encountered conflicts was measured instead.
Solving time and number of conflicts are approximately proportional, at least up to a certain limit; for very large solving times, the BCP speed of solvers usually drops because of the increasing number of learned clauses.

Using the modified solver, some industrial SAT problems used in the 2005 SAT competition and the 2006 SAT Race \cite{SATRACE:2006} were solved a large number of times with random initializations, and the number of encountered conflicts was recorded for each initialization.
Let the number of conflicts using the standard initialization for a problem be $\kappa_0$, then a random initialization will yield a number of conflicts that is a percentage of $\kappa_0$.
Counting and charting the number of occurrences of the (rounded) percentages gives an \emph{initialization histogram} (e.g.\ Figure~\ref{fig:stric-bmc-ibm-10-1000,0,1.fig}).
In the caption of the following histogram figures, $N_v$ and $N_c$ are the numbers of variables and clauses of the problem, $T_0$ is the solving time with standard initialization on a 2.4 GHz Pentium 4 PC; the number of samples taken and the range of the random initialization numbers are also given.

\begin{figure}
\centering
\includegraphics[width=0.5\textwidth]{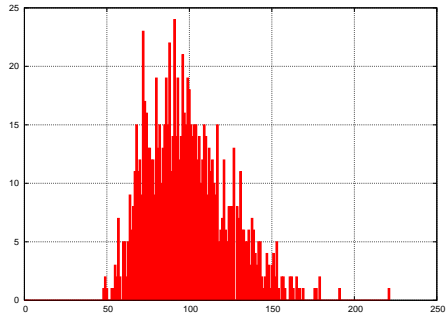}
\caption{{\ttfamily stric-bmc-ibm-10} ($N_v$ = 59056, $N_c$ = 323700, $T_0$ = 3 s, $\kappa_0$ = 4137), 1000 samples, range $[0; 1]$\label{fig:stric-bmc-ibm-10-1000,0,1.fig}}
\end{figure}

Figure~\ref{fig:stric-bmc-ibm-10-1000,0,1.fig} shows the histogram of the problem \hyphtt{stric-bmc-ibm-10} (satisfiable), using random fractional numbers between 0 and 1 for the activities, and solving 1000 times.
The random number range $[0, 1]$ was chosen arbitrarily; some experimenting with larger ranges like $[0, 1000]$ indicated little difference in the resulting histograms.
The lowest number of conflicts found was 1995 or 48\% of $\kappa_0$, the highest was 9126 conflicts (221\% of $\kappa_0$).
Better or worse initializations (that were not found by this experiment) may exist.
The histogram has a vague bell shape, with a peak near 100\%.
Only a few occurrences of very low or very high numbers of conflicts were found.
For this problem, it is clear from the chart that the number of conflicts can be reduced to at least half of $\kappa_0$, if the algorithm can compute the required initialization from the CNF.

\begin{figure}
\centering
\includegraphics[width=0.5\textwidth]{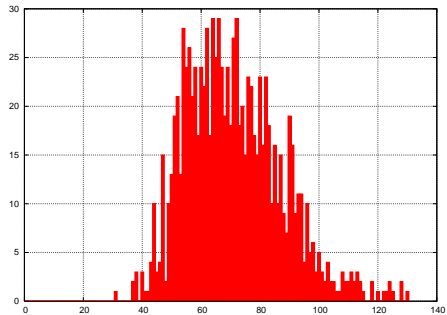}
\caption{{\ttfamily stric-bmc-ibm-10} reordered ($T_0 = 9 s, \kappa_0 = 7054$), 1000 samples, range $[0, 1]$\label{fig:stric-bmc-ibm-10.s3,1000,0,1.fig}}
\end{figure}

It has been found in the SAT competitions that \emph{reordering} the CNF of a SAT problem usually affects the solving time negatively.
Reordering changes the order of the clauses and renames and inverts variables, which does not change the satisfiability of the problem but changes the progression of the DPLL search.
Next it will be investigated if and to what degree activity initialization may be used to counter these effects.
\hyphtt{stric-bmc-ibm-10} was reordered with \textsc{reorder.c} \cite{Simon:2002} (random seed was 3); the reordered problem takes 9 seconds (300\% of original) and $\kappa_0 = 7054$ conflicts (170\% of original) to solve.
When tested with random initializations (Figure~\ref{fig:stric-bmc-ibm-10.s3,1000,0,1.fig}), the lowest number of conflicts found was 2207 (31\% $\kappa_0$) vs.\ 1995 for the original problem, the highest was 9165 (130\%  $\kappa_0$) vs.\ 9126 originally.

Since the shapes of the curves in Figure~\ref{fig:stric-bmc-ibm-10-1000,0,1.fig} and Figure~\ref{fig:stric-bmc-ibm-10.s3,1000,0,1.fig} resemble each other and both could be solved with similar minimum and maximum numbers of conflicts, it is conjectured that the large discrepancy in $\kappa_0$ for the problems is at least in part due to different initial decisions, which could be remedied by computing an optimized initialization which identifies good decisions by the CNF structure rather than by more or less arbitrary variable indexes.
An effect of reordering that can not be affected by optimized initialization is due to the fact that \textsc{MiniSAT} always assigns decision variables the value \emph{false} first.
If variables have been inverted, opposite branches of the respective decisions will be explored in the original and reordered CNFs, possibly leading to a very different course of the search.

\begin{figure}
\centering
\includegraphics[width=0.5\textwidth]{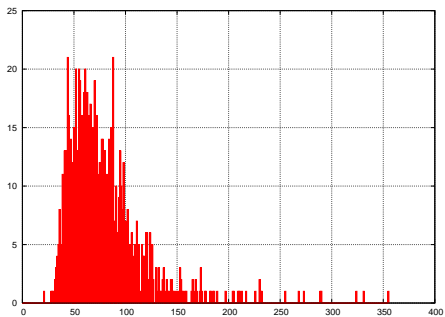}
\caption{{\ttfamily velev-sss-1.0-cl} ($N_v = 1453, N_c = 12531, T_0 = 1 s, \kappa_0 = 15211$), 1000 samples, range $[0, 1]$\label{fig:velev-sss-1.0-cl,1000,0,1.fig}}
\end{figure}

The range over which the number of conflicts can be varied with initialization seems to be highly dependent on the problem.
Figure~\ref{fig:velev-sss-1.0-cl,1000,0,1.fig} shows the histogram of problem \hyphtt{velev-sss-1.0-cl} (unsatisfiable), which has a $\kappa_0$ three times that of \hyphtt{stric-bmc-ibm-10} but takes less than half as much time to solve, presumably because it has only 1/25th as many clauses.
Compared to Figure~\ref{fig:stric-bmc-ibm-10-1000,0,1.fig} the histogram has a more spread-out appearance.
The number of conflicts ranges from 3262 (21\% $\kappa_0$) to 53983 (355\% $\kappa_0$), with a broad clustering near 50\%.
Again, extremely low or high numbers of conflicts are sparse, most values are clustered around a peak.

\begin{figure}
\centering
\includegraphics[width=0.5\textwidth]{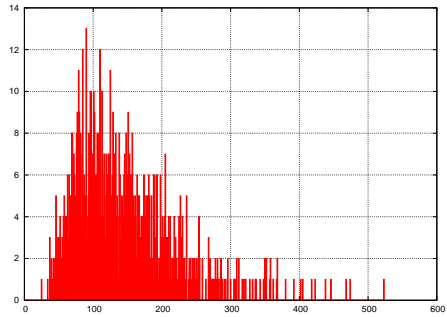}
\caption{{\ttfamily velev-sss-1.0-cl} reordered ($T_0 = 3 s, \kappa_0 = 30791$), 1000 samples, range $[0, 1]$\label{fig:velev-sss-1.0-cl-sh2,1000,0,1.fig}}
\end{figure}

Reordering \hyphtt{velev-sss-1.0-cl} (with random seed 2) yields a problem that has a $\kappa_0$ twice that of the original and takes 3 times as long to solve.
The lowest number of conflicts found (Figure~\ref{fig:velev-sss-1.0-cl-sh2,1000,0,1.fig}) was 7649 (25\%  $\kappa_0$) vs.\ 3262 for the original problem, the highest was 161093 (523\% $\kappa_0$) vs.\ 53983 originally.
For this problem, reordering made solving much harder and increased the spread of the histogram.
Yet, a good initialization can reduce the number of conflicts to half that of the original, unreordered problem.

\begin{figure}
\centering
\includegraphics[width=0.5\textwidth]{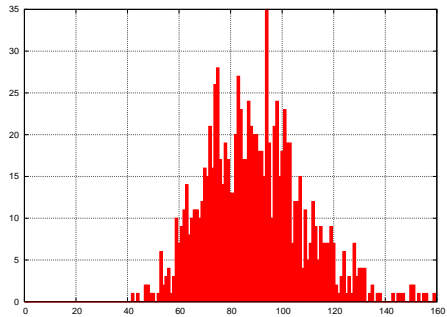}
\caption{{\ttfamily manol-pipe-g7n} ($N_v = 23936, N_c = 70492, T_0 = 10 s, \kappa_0 = 25163$), 1000 samples, range $[0, 1]$\label{fig:manol-pipe-g7n}}
\end{figure}

For \hyphtt{manol-pipe-g7n} (Figure~\ref{fig:manol-pipe-g7n}) random initialization found a range from 10645 conflicts (42\% $\kappa_0$) to 40128 conflicts (159\% $\kappa_0$).
The corresponding solving times were $3s$ and $19s$; it can be seen that the number of conflicts and the solving time are not fully proportional.

\begin{figure}
\centering
\includegraphics[width=0.5\textwidth]{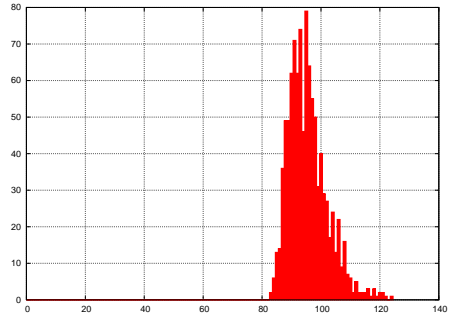}
\caption{{\ttfamily velev-eng-uns-1.0-04a} ($N_v = 7000, N_c = 67586, T_0 = 97 s, \kappa_0 = 105931$), 1000 samples, range $[0, 1]$\label{fig:velev-eng-uns-1.0-04a,1000,0,1.fig}}
\end{figure}

$\kappa_0$ may be more or less near the optimal initialization.
For problem \hyphtt{velev-sss-1.0-cl} (Figure~\ref{fig:velev-sss-1.0-cl,1000,0,1.fig}) there were clearly many better initializations.
Figure~\ref{fig:velev-eng-uns-1.0-04a,1000,0,1.fig} shows the histogram of problem \hyphtt{velev-eng-uns-1.0-04a} (unsatisfiable), which is much harder than the previous problems.
The best initialization found resulted in 87544 (83\% $\kappa_0$), the worst had 130965 conflicts (124\% $\kappa_0$).
The spread here is much lower than for the other problems.
There are better initializations than the standard for this problem, but there seems to be less room for improvement than for the other problems.

\begin{figure}
\centering
\includegraphics[width=0.5\textwidth]{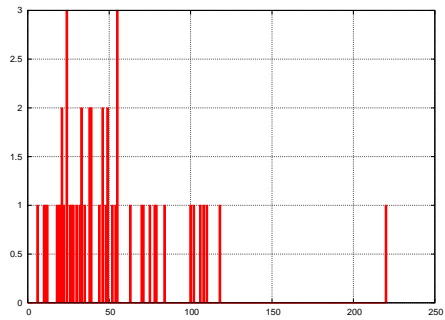}
\caption{{\ttfamily simon-mixed-s02bis-01} ($N_v = 2424, N_c = 14812, T_0 = 882 s, \kappa_0 = 2238242$), 50 samples, range $[0, 1]$\label{fig:thanner-simon-mixed-s02bis-01,50,0,1.fig}}
\end{figure}

Figure~\ref{fig:thanner-simon-mixed-s02bis-01,50,0,1.fig} shows the histogram of problem \hyphtt{simon-mixed-s02bis-01} (satisfiable).
It is even harder than \hyphtt{velev-sss-1.0-cl} despite having much fewer variables and clauses.
Only 50 samples were taken for this problem due to the large solving time (on a 3.2 GHz Pentium 4).
The lowest number of conflicts found was only 141231 (6\% $\kappa_0$) with a solving time of 37 s (4\% $T_0$).
Obviously, for this hard problem a good initialization could make the difference between time-out and success.

It should be noted that the initial decisions are changed only in their order, not in their polarity, i.e.\ the first assigned value is always \emph{false}.
Therefore, this is not equivalent to a ``guess'' at a model for satisfiable problems.

\section{Genetic Programming}

In Section~\ref{random.init.seq} it was experimentally investigated how the initial activities influence the DPLL SAT solving process.
It was found that it is possible to significantly reduce the number of conflicts if the initialization is good; what makes an initialization beneficial and how to compute it is unknown.
Good random initializations occurred relatively rarely, and may be hard to compute.
Since SAT is NP-complete, it is likely that this ``DPLL initialization problem", which is similar to the problem of finding an optimal variable order, is hard as well.
An exact algorithm is likely to be of exponential complexity and may take longer than the actual SAT solving.
A heuristic approach is more promising.
Instead of manually designing and testing heuristics one by one, the approach in this work is modeled on the procedure of \emph{evolutionary algorithms}.

First, a solution template is designed which should, ideally, be able to describe all possible solutions (including all optimal ones) of the problem at hand; this is the \emph{phenotype}.
The solutions have to be encodable in a form that allows the application of \emph{evolutionary operators} like \emph{crossover} and \emph{mutation}; this is the \emph{genotype}.
A \emph{population} of randomly generated solutions, the \emph{individuals}, is created.
Using a number of \emph{fitness cases} (concrete instances of the problem) the individuals are evaluated and assigned a \emph{fitness}, a number quantifying their success, by the \emph{fitness measure}.
More fit individuals are more likely to have offspring for the next \emph{generation}.
Over the course of many generations the average fitness increases.
Evolutionary algorithms are capable of finding very good, often non-obvious solutions.

\begin{figure}
\centering
\includegraphics[width=0.4\textwidth]{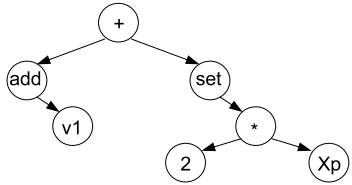}
\caption{Example of a GP parse tree\label{fig:example-tree}}
\end{figure}

\emph{Genetic programming (GP)} \cite{Koza:1994} uses \textsc{Lisp} \emph{S-expressions} (parse trees, see Figure~\ref{fig:example-tree}) composed of \emph{terminal-} and \emph{function-nodes} as the genotype.
The phenotype are complex computer programs which may contain conditional operators and memory operations.
\emph{Terminals} provide problem-specific information or constant values, while \emph{functions} take a number of arguments and return a result (they may also have side effects).
To ensure that any arrangement of nodes is valid, all terminals, function arguments and return values must have the same data type (\emph{closure requirement}).
The evolutionary operators \emph{crossover} and \emph{mutation} work on trees by exchanging or modifying nodes and subtrees.

This work improves on the concepts introduced in \cite{Kibria:2006}.
The initialization algorithm in \cite{Kibria:2006} can describe a subset of the solutions possible in this work.

\subsection{\label{Initialization.Algorithm.sec}Initialization Algorithm}

The initialization algorithm has to compute the initial activity value (a floating-point number) for each variable from the CNF.
Some information that can be easily extracted from the CNF includes the number of variables and clauses and how often literals occur.
The influence of a variable assignment depends not only on how often the literal occurs, but also with which other literals and in which polarity it occurs in the clauses.
The heuristic to be evolved should have sufficient information available to find a good initialization, but should not require extensive preprocessing to extract more complex information from the CNF.

As a compromise, it is assumed that a single iteration over all clauses that contain the variable whose activity is computed can gather sufficient information to compute an adequate result, if not an optimal one.
All literals occurring together with the variable's literal in each clause can be examined, and the activity can be modified accordingly.
The pseudocode of this algorithm template (in \textsc{C}-style notation) is:

\begin{verbatim}
double computeActivity (Variable X) {
  double a0 = 0.0, v1 = 0.0, v2 = 0.0;
  PRE_LOOP_CODE;
  for (all clauses C containing X) {
    for (all literals L in C, except the literal of X) {
      IN_LOOP_CODE;
    }
  }
  POST_LOOP_CODE;
  return a0;
}
\end{verbatim} 

The function \hyphtt{computeActivity} computes the initial activity of SAT variable $X$ and returns it.
It is called once each for all variables of the CNF.
Inside the function the variable \hyphtt{a0}, which is set to 0 at the beginning, stores the activity.
This variable's value can be modified with some operators that will be made available to GP.
Additionally there are two variables \hyphtt{v1} and \hyphtt{v2} for storing and computing auxiliary results.

The placeholders \hyphtt{PRE\_LOOP\_CODE}, \hyphtt{IN\_LOOP\_CODE} and \hyphtt{POST\_LOOP\_CODE} stand for the program fragments that will be evolved with GP.
They will contain expressions with and without side effects (e.g.\ changing one of the variables).
If a program fragment contains only expressions without side effects, it is equivalent to empty code.
In all fragments some basic information (in the form of terminals) is available, e.g.\ how often $X$ occurs.
For \hyphtt{IN\_LOOP\_CODE} some additional information about the clause $C$ and the literal $L$ is made available.

\subsection{Terminal and Function Sets}

The three program fragments of the initialization algorithm have different sets of available terminals.
\hyphtt{PRE\_LOOP\_CODE} and \hyphtt{POST\_LOOP\_CODE} are outside the loop, so they do not have access to terminals describing the clause $C$ and the literal $L$.
Table~\ref{terminals.table} shows all available terminals.

\begin{table}
\caption{Terminal set\label{terminals.table}}
\begin{center}
\begin{tabular}{lr}
\hline\noalign{\smallskip}
Terminal & Meaning \\ 
\noalign{\smallskip}
\hline
\noalign{\smallskip}
& \textbf{Available in all program fragments:} \\
\hyphtt{xn, xp, xc} & $\#$ of negative/positive/total literals of $X$ in the CNF \\ 
\hyphtt{nv, nc} & $\#$ of variables/clauses of the CNF \\ 
\hyphtt{0, 1, 2, 3, 4} & The numerical constants 0, 1, 2, 3, 4 \\
\hyphtt{a0, v0, v1} & Current values of variables \hyphtt{a0}, \hyphtt{v1} and \hyphtt{v2} \\
\hline
& \textbf{Only for \hyphtt{IN\_LOOP\_CODE}:} \\
\hyphtt{ln, lp, lc} & $\#$ of negative/positive/total literals of variable of $L$ in the CNF \\ 
\hyphtt{cs} & $\#$ of literals in current clause $C$ \\ 
\hyphtt{xs, ls} & Polarity of $X$ in clause $C$ / current literal $L$ (0 negative, 1 positive) \\
\hyphtt{ic, il} & Index of clause $C$ / literal $L$ (0 $\leq$ \hyphtt{ic} $<$ \hyphtt{xc}, 0 $\leq$ \hyphtt{il} $<$ \hyphtt{cs} $-$ 1)\\
\hline
\end{tabular}
\end{center}
\end{table}

A number of functions with one, two or three arguments are available (Table~\ref{functions.table}) for all program fragments.
The function return values as well as the currently computed activity are restricted to the (arbitrary) range $[-10^6, 10^6]$.
If a value is lower than $-10^6$ or higher than $10^6$, it is set to the respective maximum negative or positive limit.
The arithmetic division operation is implemented in such a way that division by very small values or zero returns the positive or negative limit.

\begin{table}
\caption{Function set\label{functions.table}}
\begin{center}
\begin{tabular}{lr|lr}
\hline\noalign{\smallskip}
Function & Return value & Function & Return value \\
\noalign{\smallskip}
\hline
\noalign{\smallskip}
\hyphtt{add(v)} & $a0 := a0 + v$, return new $a0$  & \hyphtt{sqrt(v)} & $v<0: -1, v \geq 0: \sqrt{v}$ \\
\hyphtt{sub(v)} & $a0 := a0 - v$, return new $a0$  & \hyphtt{abs(v)} & $|v|$ \\
\hyphtt{mul(v)} & $a0 := a0 \times v$, return new $a0$  & \hyphtt{progn2(x,y)} & Evaluate x and y, return y \\
\hyphtt{div(v)} & $a0 := a0 \div v$, return new $a0$  & \hyphtt{min(x,y)} & min(x,y) \\
\hyphtt{set(v)} & $a0 := v$, return new $a0$   & \hyphtt{max(x,y)} & max(x,y) \\
\hyphtt{setv1(v)} & $v1 := v$, return new $v1$ & \hyphtt{and(x,y)} & $(x>0) \wedge (y>0): 1$, else 0 \\
\hyphtt{setv2(v)} & $v2 := v$, return new $v2$ & \hyphtt{or(x,y)} & $(x>0) \vee (y>0): 1$, else 0 \\
\hyphtt{inv(v)}   &  $\frac{1}{v}$ & \hyphtt{xor(x,y)} & $(x>0 \wedge y \leq 0) \vee$ \\
\hyphtt{neg(v)}   &  $-v$ & & $(x \leq 0 \wedge y>0):1$, else 0 \\
\hyphtt{exp(v)}   &  $e^v$ & \hyphtt{lessthan(x,y)} & $x < y : 1$, else 0\\
\hyphtt{log(v)}   &  $v \leq 0: -10^6, v > 0: ln(v)$ & \hyphtt{x} \{\hyphtt{+,-,*,\%}\} \hyphtt{y} & Arithmetics \\
\hyphtt{sgn(v)}   &  $v<0: -1, v=0: 0,$ & \hyphtt{progn3(x,y,z)} & Evaluate x, y, z, return z \\
                  &  $v>0: 1$           & \hyphtt{if(x,y,z)} & $x>0: y, x \leq 0: z$ \\
\hline
\end{tabular}
\end{center}
\end{table}

\subsection{Fitness Measure}

The progress of evolution in GP is controlled by the fitness measure.
Individuals are assigned a fitness value, which is a single number signifying the quality of the solution.
The \emph{lower} the fitness value, the better the solution.
The fitness cases for GP will be one or more SAT problems.
Each individual will be used to compute an initialization for the problems; solving yields the number of conflicts encountered.
In the implementation in this work no time-out is possible; the solver runs until the problem is solved.

The simplest fitness measure is to sum up the number of conflicts for all problems.
This would drive evolution toward an algorithm that solves the problems with the lowest total number of conflicts.
Should two individuals result in the same number of conflicts, they would be assigned equal fitness, even though they may improve the solving of each single problem to different degrees.
In such a case, the solution that improves harder problems more than easier ones is preferable.
This can be achieved by summing up the squares of the numbers of conflicts, which prefers low numbers to high numbers disproportionally.
Also returned by the SAT solver is the number of decisions made.
If two individuals achieve the same number of conflicts, the one that required a lower number of decisions should be preferred.
When the numbers of conflicts and decisions achieved by two individuals is equal, the trees that represent the program fragments may differ in complexity.
In such a case, the individual with the most compact trees is to be preferred.
A measure for the complexity of a tree is the length, i.e.\ the number of nodes it has.

Let $F_k$ be the fitness value of individual $k$, $N_p$ the number of fitness cases, $c_{k,i}$ the number of conflicts on solving problem $i$ with individual $k$ and $d_{k,i}$ the number of decisions; the total number of nodes in all trees of the individual $k$ is $l_k$.
Then $F_k$ is computed according to Equation~\ref{fitness.eq}.

\begin{equation}
\label{fitness.eq}
	F_k = \sqrt{\sum_{i=1}^{N_p} (c_{k,i} + \frac{d_{k,i}}{1000})^2} + \frac{l_k}{1000}
\end{equation}

The main influence on the fitness should be the number of conflicts, while the number of decisions and the length of the trees are of secondary importance.
Therefore the latter are divided by a factor 1000.
To scale back the sum of the squares, which can become very large, the square root is taken of it.

\section{Experimental Results}

\subsection{Implementation}

The SAT solver \textsc{MiniSAT v1.14} \cite{Een:2005} was used in this work, and the GP functionality was implemented using the \textsc{gpc++} library \cite{gpcpp}.
After a SAT problem is read from disk, the solver preprocesses it by applying BCP.
The activity initialization algorithm operates on the preprocessed CNF.

The initialization algorithm as described in Section~\ref{Initialization.Algorithm.sec} computes the activity for each variable one by one.
For various reasons it was simpler to implement the algorithm in a modified form which iterates over all clauses only once and computes all activities in one go.
This is achieved by considering all combinations of the variable $X$ and literal $L$ in each clause and running the \hyphtt{IN\_LOOP\_CODE} fragment for each combination.
The order in which the CNF's clauses are stored may lead to different activities being computed, depending on the initialization algorithm.
No measures were taken to preserve the original order of the clauses in the file.
\textsc{MiniSAT} stores binary (two-literal) clauses in the \emph{watcher lists}, from which they have to be extracted first.
The initialization algorithm iterates over the binary clauses first.

For all experiments the default parameters of the GP library were used: crossover probability 95\%, creation probability 2\%, creation type \emph{Ramped Half and Half}, maximum depth for creation was 6, maximum depth for crossover was 17.
The selection type was \emph{tournament selection}, with a tournament size of 10.
\emph{Steady state GP} was turned on, \emph{demetic grouping} was off.
Mutation probability was 0\%.
GP uses population sizes (the number of individuals) ranging from 1000 to 16000, depending on the problem \cite{Koza:1994}.
In this work, due to the large computation times, the population size in all experiments was 1000.
The usual number of generations in GP is ca.\ 21 to 51; in this work, it was 5.

\subsection{Results}

The three SAT problems \hyphtt{stric-bmc-ibm-10} (problem ``S'' henceforth), \hyphtt{velev-sss-1.0-cl} (``V'') and \hyphtt{manol-pipe-g7n} (``M''), whose initialization histograms can be seen in Section~\ref{random.init.seq}, will be used as fitness cases for GP.

The best result found in the precursor work \cite{Kibria:2006}, using 9 fitness cases, was equivalent to \hyphtt{IN\_LOOP\_CODE} = \{\hyphtt{add(exp(-lc)-lp)}\} (it always computes a negative activity).
It was capable of improving the solving of non-fitness-case problems noticably.


First it was attempted to find an initialization algorithm that can improve the solving of problem S alone.
With only one fitness case, it is likely that any solution found (especially if it is  extremely good) is specialized on that problem, with much decreased efficiency for other problems.
On the other hand, the range of the number of conflicts that results from GP can be compared to random initialization, possibly yielding cues on how reliable the results of the latter are.
The best solution for S was found in generation 0 (the first one, containing randomly generated individuals).
The individual's program fragments were:

\begin{verbatim}
 PRE_LOOP_CODE = neg (4)
  IN_LOOP_CODE = if (sub (xp), cs, inv (4))
POST_LOOP_CODE = exp (neg (xc))
\end{verbatim} 

Using this algorithm, the number of conflicts for solving the problem was 464 (11\% $\kappa_0$) and the number of decisions was 9231.
There are a total of 11 nodes in the trees, which (using Equation~\ref{fitness.eq}) results in a fitness of 473.242 for the individual.

\hyphtt{PRE\_LOOP\_CODE} and \hyphtt{POST\_LOOP\_CODE} only contain functions without side effects, so they are equivalent to empty code.
\hyphtt{IN\_LOOP\_CODE} contains the activity-modifying expression \hyphtt{sub(xp)}.
This operation subtracts the number of positive occurrences of the variable $X$ once for every literal (minus 1) in clauses that contain $X$.
The resulting activities of all variables will therefore be negative.
In the course of evolution the remaining non-functional expressions in the trees were eliminated, until the \hyphtt{PRE-} and \hyphtt{POST}-trees contained only one terminal node and the \hyphtt{IN}-tree contained only \hyphtt{sub(xp)}, but no lower number of conflicts could be achieved.
Compared to the best result found by random initialization (1995 conflicts, 48\% of $\kappa_0$), the result found by GP (464 conflicts) is a large improvement.

\begin{table}
\caption{\hyphtt{stric-bmc-ibm-10}: $\#$ conflicts with GP initializations\label{stric10conflicts.table}}
\begin{center}
\begin{tabular}{|c|c||c|c||c|c|}
\hline
orig. & reord. & $c_{orig,best}$ & $c_{orig,worst}$ & $c_{reord,best}$ & $c_{reord,worst}$ \\
\hline
\hline
$\surd$ & -       & 464 (11\% $\kappa_0$) & 9248 (224\% $\kappa_0$) & - & - \\ \hline
 -      & $\surd$ & -                     & -                       & 2145 (30\% $\kappa_0$) & 9305 (132\% $\kappa_0$) \\ \hline
$\surd$ & $\surd$ & 1239 (30\% $\kappa_0$) & 7445 (180\% $\kappa_0$) & 3014 (43\% $\kappa_0$) & 7329 (104\% $\kappa_0$) \\ \hline
\end{tabular}
\end{center}
\end{table}

Table~\ref{stric10conflicts.table} shows the best and worst results when using S alone, the reordered S or both together as fitness cases.
It can be seen that better initializations could be found when using a problem alone.
GP found a slightly better initialization than the random procedure for the reordered problem.
Table~\ref{stric10individuals.table} shows the programs corresponding to the results of these experiments (if no code is given for a fragment, it is empty).
The programs seem to have little in common.

\begin{table}
\caption{\hyphtt{stric-bmc-ibm-10}: best and worst GP individuals\label{stric10individuals.table}}
\begin{center}
\begin{tabular}{|c|c||l|l|}
\hline
orig. & reord. & Best & Worst \\
\hline
\hline
$\surd$ & -       & \hyphtt{IN: sub(xp)}       & \hyphtt{POST: add(nc+3)}      \\ \hline
 -      & $\surd$ & \hyphtt{POST: add(exp(1))} & \hyphtt{IN: set(min(xn,xp))}  \\ \hline
$\surd$ & $\surd$ & \hyphtt{PRE: add(nc)}      & \hyphtt{IN: set(xc)} \\ 
        &         & \hyphtt{IN: sub(nv)}       & \hyphtt{POST: div(xp)} \\ 
        &         & \hyphtt{POST: sub(xn), mul(2), sub(xc)} & \\ \hline
\end{tabular}
\end{center}
\end{table}

More experiments were made with S, V and M in several combinations; all three together was run twice.
Table~\ref{allproblems.table} shows the best (indexed $b$) and worst results (indexed $w$), in \% of $\kappa_0$, for each problem alone in each experiment and the average.
Table~\ref{allproblems.indiv.table} contains the resulting programs, which show no obvious structure.
For V and M, GP did not find initializations as good as the random procedure.
When using several problems, the best results for each single problem seem to get worse (with exceptions), but the average improvement is still good.
On average, solving the fitness cases with initialization takes less than half the number of conflicts as without.

\begin{table}
\caption{All problems: \% of $\kappa_0$ with GP initializations\label{allproblems.table}}
\begin{center}
\begin{tabular}{|c||c|c||c|c||c|c||c|c|}
\hline
Combination & $S_b$ & $S_w$ & $V_b$ & $V_w$ & $M_b$ & $M_w$ & $Sum_b$ & $Sum_w$ \\ \hline \hline
S           & 11    & 224   & -     & -     & -     & -     & -       & -       \\ \hline
V           & -     & -     & 23    & 583   & -     & -     & -       & -       \\ \hline
M           & -     & -     & -     & -     & 50    & 145   & -       & -       \\ \hline
SV          & 73    & 75    & 29    & 1079  & -     & -     & 38      & 864     \\ \hline
SVM1        & 9     & 98    & 39    & 495   & 52    & 106   & 43      & 239     \\ \hline
SVM2        & 62    & 74    & 38    & 399   & 52    & 111   & 48      & 206     \\ \hline
\end{tabular}
\end{center}
\end{table}

\begin{table}
\caption{All problems: best and worst GP individuals\label{allproblems.indiv.table}}
\begin{center}
\begin{tabular}{|c||l|l|}
\hline
Test      & Best                       & Worst                    \\ \hline \hline
S         & \hyphtt{IN: sub(xp)}       & \hyphtt{POST: add(nc+3)} \\ \hline

V         & \hyphtt{PRE: div(lessthan(2,xn))} & \hyphtt{IN: add(and(ls,xs))}       \\ 
          & \hyphtt{POST: div(2)}             &         \\ \hline

M         & \hyphtt{IN: add(ln+xs+1)}         &  \hyphtt{POST: set(nv)}      \\ \hline

SV        & \hyphtt{IN: set(lp)}              &  \hyphtt{PRE: add(sgn(set(xp)+nc-3))}      \\
          & \hyphtt{POST: add(1)}             &  \hyphtt{IN: setv1(add(setv1(xs))),setv1(setv2(xc))} \\
          &                                   &  \texttt{POST: if(xor(v2,a0)\%add(v1),0,sub(xc))} \\ \hline

SVM1      & \hyphtt{PRE: set(xn)}             & \hyphtt{IN: setv2(set(xp)),}       \\ 
          & \hyphtt{IN: div(lp)}              & \hyphtt{    if(lessthan(ln,v2),0,mul(il))} \\ 
          & \hyphtt{POST: add(1)}             &  \\ \hline

SVM2      & \hyphtt{IN: set(3),div(2)}        & \hyphtt{IN: div(sub(1))}       \\ \hline

\end{tabular}
\end{center}
\end{table}

\subsection{Result used in SAT-Race 2006}

Additionally to the base algorithm used in the experiments described before, a \emph{normalization} phase was added:
it searches for the largest absolute initialization value and divides all activities by it, so that all activities are between $-1$ and $1$.
This was done because for many problems (especially large, industrial ones) the computed raw activities had very large values, with unknown effect on the decision heuristic.
With normalization only the first few decisions should be influenced.
Using normalization did not seem the range of influence (factor 10 better or worse) of the initialization.

Many more GP runs were executed, most of which yielded strongly varying results.
To further filter out flukes, a validation test was done using the problems from the first qualification round of SAT-Race 2006.
The best result found was:

\begin{verbatim}
 PRE_LOOP_CODE = {}
  IN_LOOP_CODE = add (lc)
POST_LOOP_CODE = {}
\end{verbatim} 

\section{Conclusion}

In this work it was found that the initial values of the activities in VSIDS-like decision heuristics, e.g.\ in \textsc{MiniSAT}, have a strong effect on the total number of conflicts encountered to solve SAT problems.
Good initializations can reduce that number by a factor 10 or more, and improve solving time accordingly, while ``bad'' ones can increase the number of conflicts by a factor 10.
Using experiments with random values, it was found that the range of effect of initialization depends on the problem.
Reordered problems were found to be affected similarly, but not necessarily equally by initialization as the original ones.

With little indication of the requirements of an algorithm to compute beneficial activities from any CNF, the approach taken was to design an ad hoc algorithm template (based on a precursor work) which has some information about the CNF and a set of operations.
Using genetic programming, randomly generated solutions could be evolved and optimized driven by a fitness measure favoring lower numbers of conflicts, with real, industrial SAT problems as fitness cases.
The evolved solutions could compute beneficial activities for one or more problems, but the underlying principle could not be discerned.

To the author's knowledge, the initialization of VSIDS activities has not been researched thoroughly.
The experiments in this work seem to indicate that the potential gain in solving speed is large, about one order of a magnitude, without requiring any changes to the other parts of the SAT solver.
It is unknown if the algorithm template presented in this work is capable of representing an algorithm that can compute beneficient activities for the general case.


\end{document}